\documentclass[12pt,a4paper]{article}

\usepackage[T1]{fontenc}
\usepackage[utf8]{inputenc}
\usepackage{tgtermes}
\usepackage[a4paper,left=2.4cm,right=2.4cm,top=2.6cm,bottom=2.6cm]{geometry}
\usepackage{amsmath}
\usepackage{graphicx}
\graphicspath{{figs/}{./}}
\usepackage[table]{xcolor}
\usepackage{array}
\usepackage{tabularx}
\usepackage{booktabs}
\usepackage[labelsep=colon,font={small},labelfont={}]{caption}
\usepackage{enumitem}
\usepackage{float}
\usepackage{titlesec}
\usepackage{textcomp}
\usepackage{listings}
\definecolor{codebg}{HTML}{F6F6F4}
\definecolor{codecmt}{HTML}{6A737D}
\definecolor{codekw}{HTML}{2B4EA2}
\definecolor{linkblue}{HTML}{2B4EA2}
\definecolor{gridgrey}{HTML}{BBBBBB}
\usepackage{xurl}
\usepackage[colorlinks=true,urlcolor=linkblue,linkcolor=black,citecolor=black]{hyperref}

\usepackage{lineno}

\titleformat{\section}{\normalfont\bfseries}{}{0pt}{\MakeUppercase}
\titleformat{\subsection}{\normalfont\bfseries}{}{0pt}{}
\titleformat{\subsubsection}{\normalfont\bfseries\itshape}{}{0pt}{}
\titlespacing*{\section}{0pt}{1.5em}{0.7em}
\titlespacing*{\subsection}{0pt}{1.15em}{0.5em}
\titlespacing*{\subsubsection}{0pt}{1.1em}{0.5em}

\arrayrulecolor{gridgrey}

\newcommand{\mcu}{m\textsuperscript{3}}

\begin{document}
% \linenumbers
\pagestyle{plain}

% ---------------- Title block ----------------
\begin{center}
{\fontsize{14}{17}\selectfont\bfseries Which Site, and When: A Free-Satellite-Data Test of Himalayan Glacial Lake Bursts, Landslides, and Ice Floods\par}
\vspace{14pt}
{\fontsize{12}{15}\selectfont Matthew Kahn\textsuperscript{1}, Milan Arjel\textsuperscript{2}, Nirmala Adhikari\textsuperscript{3}, Mingmar Sherpa\textsuperscript{4}, James Pope\textsuperscript{5,*}\par}
\vspace{12pt}
{\fontsize{12}{15}\selectfont \textsuperscript{1}Applied Economics and Management, Cornell University,\\ New York, United States of America\par}
\vspace{6pt}
{\fontsize{12}{15}\selectfont \textsuperscript{2}Department of Electronics and Computer Engineering, Purwanchal Campus, Institute of Engineering,\\ Tribhuvan University, Dharan, Nepal\par}
\vspace{6pt}
{\fontsize{12}{15}\selectfont \textsuperscript{3}Department of Physics, University of Alabama at Birmingham,\\ Alabama, United States of America\par}
\vspace{6pt}
{\fontsize{12}{15}\selectfont \textsuperscript{4}Department of Biomedical and Biological Sciences, Cornell University,\\ New York, United States of America\par}
\vspace{6pt}
{\fontsize{12}{15}\selectfont \textsuperscript{5}School of Engineering Mathematics and Technology, University of Bristol,\\ Bristol, United Kingdom\par}
\vspace{10pt}
{\fontsize{12}{15}\selectfont \textsuperscript{*}Corresponding author: James Pope \{jp16127@bristol.ac.uk\}\par}
\end{center}
\vspace{10pt}

\section{Abstract}

Two free satellite signals carry real information about glacial-lake outburst risk in the Nepal Himalaya: radar interferometry sees a moraine dam slowly sagging, and satellite weather marks the weeks when a primed lake is under stress. A companion feasibility study found that deformation indicates which lake is destabilizing and weather indicates when it is at risk, but proposed no predictive model. To address this gap, we propose and evaluate models that predict which site is susceptible and when a trigger arrives. We test three related hazards on free data alone: large moraine- and ice-dammed bursts, rainfall-triggered landslides, and smaller floods from ponds on and around a glacier. Each hazard gets two questions, never blended. Using 589 dated outbursts from HMAGLOFDB and several thousand catalogued landslides, we match each event against similar but unfailed sites, and hold every model to a strong simple baseline under spatial cross-validation that withholds whole map tiles, so no model succeeds by recognising a trained-on neighbourhood. Antecedent weather times the trigger at ROC 0.73 for big bursts, 0.83 for landslides, and 0.82 for small floods. Terrain ranks susceptibility only in part: scored naively it appears near 0.9, largely because catalogued failures cluster in wetter ranges; matched against comparable nearby sites the honest figures are 0.76, 0.71, and 0.54 (no better than chance). The burst signal holds within single regions, reaching 0.89 in Nepal alone. Five deep-learning models do not decisively beat a simple gradient-boosted baseline. Three score marginally higher on landslides, a hint too small to confirm. For the lake hazards the baseline wins outright, reproduced by a three-rule decision tree on ruggedness and monsoon rainfall. We close with a ranked Nepal watchlist, a prioritisation aid, not a prediction, and note where free data reaches its limits.

\noindent\textbf{Keywords:} Glacial lake outburst flood; GLOF; landslide; Nepal Himalaya; hazard prediction; susceptibility; interpretable machine learning; free satellite data
\section{Introduction}

A glacial lake outburst flood, or GLOF, is the sudden release of water impounded behind a natural dam of glacier ice or glacial sediment. In the Nepal Himalaya the dams of concern are mostly moraines: ridges of unconsolidated rock and till forced upward by a glacier and left behind as it retreats, often containing a buried core of relict ice. When such a dam fails, the lake it held can empty in hours, sending a flood of water, sediment, and boulders into the valleys below and destroying homes, bridges, farmland, and hydropower installations far downstream. Regional catalogs record hundreds of historical outbursts across the Hindu Kush Himalaya [21], and the glacial lakes that feed them are growing: between 1990 and 2018 the world's glacial lakes grew by roughly 53\% in number, 51\% in area, and 48\% in stored volume as the glaciers retreated [22]. On the night of 3 October 2023 the moraine-dammed South Lhonak Lake in Sikkim, on Nepal's eastern border, discharged about 50 million~\mcu{} of water after a mass of frozen lateral moraine collapsed into it [30]; the resulting flood killed at least 55 people, left dozens more missing, and destroyed the 1{,}200-megawatt Teesta III dam at Chungthang, the largest hydropower project on the river [20]. Nepal treats the risk as a national priority, and a roughly \$36 million Green Climate Fund program with the government and the United Nations Development Programme is underway to monitor and lower the most dangerous lakes [24].

Whether that growing lake volume translates into more frequent floods is genuinely contested. Two independent inventories find that the annual frequency of moraine-dam outbursts in the Himalaya has not risen, and may have fallen, even as lake area has expanded, so that the number of floods per unit of lake area has declined [25], [11]. The reconciliation is that a moraine dam fails only when a discrete trigger overwhelms it, and triggers do not scale with lake area. This distinction between the slow build-up that makes a lake vulnerable and the trigger that sets it off runs through the entire analysis below.

Nepal's Department of Hydrology and Meteorology (DHM) and the International Centre for Integrated Mountain Development (ICIMOD) maintain lake inventories, assess hazard level, and operate in-situ early-warning instrumentation on a small number of the most-watched lakes. What is yet to exist is a continuous, automated layer that observes every lake in a basin and flags those that are changing, the building of which directs the flow of this paper. The intended end product is not a siren that sounds minutes before a flood wave, which remains the province of DHM's downstream in-situ sensors, but a screening layer operating at a lead time of weeks to months: a ranked, evidence-bearing watchlist of which lakes are becoming dangerous, meant to feed the institutions that hold the mandate rather than to replace them.

The economic logic for attempting this from free satellite data is direct. The alternative to automated satellite screening is periodic manual inventory and episodic field survey, which do not scale to the thousands of glacial lakes in the Nepal Himalaya. Freely licensed satellite data carries no marginal acquisition cost, so the binding constraint on all-lakes monitoring becomes analysis rather than data purchase, which is precisely the constraint a small technical team can relieve. Against the downstream cost of a single event, of which the destruction of the Teesta hydropower cascade is one recent measure, continuous screening is inexpensive. This assessment therefore holds itself to free data throughout, tests how far that takes the problem, and quantifies where it stops.

The companion feasibility assessment [36] established which free signals carry information; it did not make a forward prediction. This paper takes that step. It runs an explicit prediction test on three related hazards at once. It keeps two questions apart, asking which site is primed and when a trigger arrives. It holds every learned model to a strong simple baseline. And it validates only on sites and regions the model never saw.

One difference of scope matters. The feasibility study looked at Nepal and modelled moraine dams alone, since ice dams barely register in the Nepal hazard. A Nepal-only, moraine-only sample is too small to test a prediction on, so we widen to all of High Mountain Asia and admit the ice-dammed and smaller-lake populations too, then transfer the result back to Nepal. What follows is that test, what it shows about the ceiling free data reaches for prediction, and a ranked Nepal watchlist built from the parts that survived it.
\section{Related Work}

\subsection{The phenomenon and the global picture}

A GLOF is a consequence of glaciers retreating from the extent they held during colder centuries. As a glacier thins and pulls back, meltwater collects in the basin behind its terminal moraine, forming a moraine-dammed lake, or on the debris-covered glacier surface as coalescing supraglacial ponds, or in a tributary valley blocked by the trunk glacier as an ice-dammed lake. Global mapping from a quarter-million satellite images shows this lake population expanding rapidly with twentieth- and twenty-first-century warming [22]. The historical record of the floods themselves is compiled in event catalogs; the High-Mountain-Asia GLOF database (HMAGLOFDB) used here records 768 events across the region, 58 of them in Nepal, with dates, dam types, and downstream impacts where known [21].

More and larger lakes need not mean more floods. Two Himalayan inventories report no credible upward trend in annual outburst frequency over recent decades, so that outbursts per unit of lake area have declined [25], [11]. The explanation these studies give is that outburst timing is set by triggers rather than by lake size, and that many of the largest and most dangerous lakes are monitored and lowered to reduce their hazard. That explanation matters here in two ways: it warns against reading lake size as a probability of failure, and it explains why a forward-looking inventory of dangerous lakes barely overlaps a catalog of lakes that have already burst.

\subsection{Dam types and failure mechanisms}

The three dam types fail by different physics, and the hazard literature treats them as separate populations rather than pooling them. A moraine dam is unconsolidated sediment, frequently with a buried dead-ice core; it fails by trigger-driven overtopping, in which an external mass movement such as an ice avalanche or rockfall drops into the lake and generates a displacement wave, or seiche, that overtops the dam crest. Once water flows over the crest it cuts a channel backward into the dam in a runaway feedback, deeper channel giving greater discharge giving faster erosion, until a breach opens. A moraine breach is generally a one-time catastrophic failure: once deeply breached the moraine cannot re-impound the lost volume. An ice dam, by contrast, typically fails by hydrostatic flotation, when the lake rises to roughly nine-tenths of the ice-dam height and the water floats and pipes beneath the ice; the glacier then flows back and reseals the dam, so the lake refills and re-drains cyclically. These recurring ice-dam floods are known as j\"okulhlaups. Supraglacial ponds drain englacially through the glacier and are small, ephemeral, and frequent.

Separating these populations is standard practice, not a convenience of this study. International hazard assessment guidance treats ice-marginal, moraine-contained, and bedrock-dammed features as distinct hazard domains [10], because susceptibility criteria, breach behaviour, and mitigation differ by dam structure; a geometric predictor meaningful for a moraine dam, such as the ratio of dam width to height, is meaningless for ice-dam flotation. The distinction also carries a prediction consistent with the data: because moraine failures are one-time and ice-dam failures are cyclical, the two have very different temporal predictability, and pooling them obscures both. This assessment therefore models moraine-dammed lakes only, a choice the Nepal data support: of the 58 Nepal events in HMAGLOFDB, 37 (64\%) are moraine-dammed and only 1 is ice-dammed, whereas ice-dammed events make up 31\% of the wider region; the remaining Nepal events are small supraglacial ponds that drain englacially. Every one of ICIMOD's 21 dangerous Nepal lakes is moraine-dammed. Ice dams are thus a negligible share of the Nepal hazard we target.

\subsection{Triggers and the ceiling on prediction}

The triggers of moraine-dam failure fall into two families. Meteorological triggers, intense rainfall, high temperature, and rapid melt, act by raising the lake level and the pressure on the dam or by delivering a surge of water. Mass-movement triggers, ice or rock avalanches and moraine collapse, act by dropping material into the lake to make the overtopping wave. The meteorological family is partly predictable: studies of the driving conditions find that antecedent precipitation accumulated over roughly a month and accumulated temperature carry the signal, with reported 95th-percentile daily thresholds near 11.7 mm of rain and 8.8 degrees Celsius [28], [9]. The mass-movement family is not predictable from weather at all, because an avalanche can occur on a clear, calm day. That unpredictability sets an intrinsic ceiling on any weather-based forecast: a model can flag the weather-driven subset of events but is blind, by construction, to the avalanche-driven remainder.

\subsection{Measuring lakes from space}

Two free satellite systems dominate glacial-lake mapping, and they are complementary. Sentinel-1 is a two-satellite constellation carrying a C-band synthetic aperture radar (SAR), an active sensor that emits its own microwave pulses and measures the returned echo, or backscatter [23]. Because it supplies its own illumination and microwaves pass through cloud, Sentinel-1 images day or night and through the monsoon overcast that blinds optical sensors, and it revisits a given location every 6 to 12 days. Calm open water is nearly a mirror to radar at these wavelengths: it reflects the pulse away from the sensor and returns very little energy, so water appears dark. Separating water from land is therefore a matter of thresholding the backscatter image, for which the standard automatic method chooses the threshold that best splits the image histogram into two classes [19]. Sentinel-2 is the companion optical constellation, imaging in 13 spectral bands from the visible to the shortwave infrared at up to 10 m resolution [5]. On cloud-free dates it gives a sharper water boundary through band-ratio indices: the Normalized Difference Water Index (NDWI), which contrasts green and near-infrared reflectance to highlight water [16], and its modified form (MNDWI), which substitutes a shortwave-infrared band to suppress built-up and vegetated background [27]. We use radar as the continuous backbone and optical as a clear-sky cross-check.

\subsection{The deformation precursor and interferometry}

Beyond area, the recent literature identifies deformation of the moraine dam itself as the more direct precursor of failure, and this is the pivot on which the present study turns. The relevant technique is interferometric SAR (InSAR), which exploits not the strength of the radar echo but its phase. In plain terms, the radar wave serves as an extraordinarily fine ruler: its wavelength is only a few centimetres, so comparing the phase of the returning wave between two passes over the same ground reveals whether the surface crept even a centimetre toward or away from the satellite in between. By differencing the phase of two radar images of the same ground taken at different times, an interferogram, one can measure ground displacement along the sensor's line of sight to a fraction of the radar wavelength. The reliability of each pixel is summarized by its coherence, the correlation between the two acquisitions, and the raw phase, which is known only modulo one wavelength, must be phase-unwrapped to recover absolute displacement. Stacking many interferograms with short temporal and spatial baselines into a time series is the small baseline subset (SBAS) method [2]. Applied to glacial dams, InSAR has measured real precursory motion: a retrospective analysis of South Lhonak Lake found significant moraine deformation before its 2023 collapse, coincident with where the moraine ultimately gave way [29], and at Imja, a Nepal Rank-I lake, fused InSAR and feature-tracking measured about 90 cm of dam subsidence over seven years [3]. For a free-data approach, the same Sentinel-1 mission that supplies the amplitude images for area mapping also supplies the phase data for InSAR, though the interferometric processing is not available inside amplitude-only cloud platforms and must be run through on-demand services [1] or open time-series packages such as LiCSBAS [17].

\subsection{Susceptibility, volume, and machine learning}

A parallel literature builds data-driven susceptibility models that rank lakes by their static propensity to fail. Across these studies the most informative inputs are lake volume, seismicity, precipitation, terrain slope, and proximity to rivers, with lake area ranking below volume; ensemble and hybrid classifiers reach areas under the receiver-operating-characteristic curve (AUC, a threshold-free measure of ranking skill where 0.5 is chance and 1.0 is perfect) near 0.83 [13]. The prominence of volume over area matters here because volume is not directly observable from a satellite image of the surface; it is estimated from area through empirical area-to-volume scaling relations calibrated on bathymetric surveys [4], or measured directly only by field bathymetry.

\subsection{Operational early warning}

Finally, the operational systems that this satellite layer is meant to feed are in-situ. Deployed GLOF early-warning stations record lake level, water temperature, hydrostatic pressure, moraine displacement, and downstream discharge, transmitting on cadences of minutes [26], and seismometers placed downstream have detected an outburst roughly five hours before its flood wave reached a village, giving genuine evacuation time [15]. These methods are the operational last mile: hardware on the ground, within the institutions' mandate, that provides the short-fuse warning a wide-area satellite screen cannot.

\subsection{What the literature implies for the design}

Read together, the prior work sets the design. Lake growth is real but only slowly raises a lake's vulnerability, and outburst frequency has not risen as lake area has grown, so growth alone should not be expected to predict which lake bursts. Timing is trigger-set, with a predictable weather family and an unpredictable mass-movement family, which argues for a hazard model that separates a lake's susceptibility from the day's trigger and accepts a ceiling on the trigger side. The signal the literature most directly ties to imminent dam failure is moraine deformation, observable by InSAR from the same free satellite already used for area. And volume, not area, is the size axis that matters for consequence. Each of these expectations became a hypothesis tested below, and the two that failed on our data, that mined area drops could label floods and that size or growth could rank burst probability, failed for reasons the literature had already hinted at.

\section{Data and Design for a Prediction Test}

\subsection{Three hazards, two questions}

The companion assessment studied moraine-dam bursts alone. The prediction test widens the scope to three related hazards that share drivers and physics, because doing so both matters operationally and repairs the small-sample problem that limited the first study. The three are as follows. The large bursts are the catastrophic events, where a lake held back either by a moraine, the ridge of loose rock and gravel a glacier bulldozes ahead of itself and abandons as it retreats, or by the body of a glacier itself, escapes in a matter of hours. The landslides are rainfall-triggered slope failures. They matter here for two reasons. They are catalogued in the thousands rather than the hundreds. And they run on the same physics, because a moraine holding back a lake is itself a slope of loose material, which fails once it is saturated and can no longer hold. The small floods come from the smaller and shallower bodies of water found on and around a glacier, ponds that sit on the ice surface, pockets of water held inside the ice, and the pools left where buried ice has melted out and the ground above it has collapsed. These are the most numerous events of the three, they are individually far less destructive, and they are the ones the area-based detector of the first study could not see, because a small lake that empties and refills within a season leaves no lasting trace in a satellite record of its surface area. Splitting the events this way follows the hazard literature, which treats dam and failure types as separate populations rather than pooling them [10]. The feasibility study modelled moraine dams alone, the dominant Nepal hazard; the prediction test additionally admits the ice-dammed bursts that form nearly a third of outbursts across the wider region [21] to gain statistical power, then transfers the result back to Nepal, where the operative lakes remain moraine-dammed.

Two questions are asked of each hazard and never blended into one number. The first is susceptibility: given a lake or a slope, is it the kind of place that fails, judged from the things about it that barely change from year to year, its shape, its steepness, its surroundings, and the climate it sits in. The second is triggering: given a place, is the trigger arriving, judged from the weather in the days and weeks beforehand. A useful way to hold the difference is that the first question is about the fuse being long or short, and the second is about whether it has been lit. Neither answer is much use alone. Knowing that a lake is fragile says nothing about whether this week is the dangerous one, and knowing that the monsoon has arrived says nothing about which of the region's thousands of lakes will be the one to go. The distinction is the one the first study drew between a lake's slow-building vulnerability and the discrete trigger that sets it off, now made the organizing axis of the experiment. Risk is the product of the two. Consequence is a third axis, held apart, exactly as in the earlier design.

\subsection{Events, comparison sites, and features}

Positive events are assembled across High Mountain Asia rather than Nepal alone, for statistical power, with the headline test being transfer back to Nepal. The lake bursts come from the High-Mountain-Asia GLOF database (HMAGLOFDB) [21]. Expanded to the full region and lightly cleaned for confidence, it supplies 589 dated outbursts, which the database's own dam-type field splits into the large bursts and the smaller floods. Of those, 426 large-burst and 83 small-flood lakes sit inside the analysis box with complete free features, and those enter the susceptibility test.

The landslides come from NASA's Global Landslide Catalog (COOLR) [33], again inside the same box. Several thousand drive the triggering test. Of these, 409 carry complete terrain features and enter the susceptibility test.

A prediction of which site is dangerous is only meaningful against a fair comparison. A burst lake must be compared against other glacial lakes that did not burst, not against random ground. A failed slope must be compared against steep slopes that held.

No single free inventory of ``lakes that did not fail'' is both complete and reachable. The comparison sites are therefore derived directly from Earth Engine. Permanent, high-elevation surface water from the Joint Research Centre's global surface-water record [35] supplies 2{,}500 candidate lakes. Steep mountain terrain from the Copernicus 30 m elevation model [7] supplies 3{,}000 candidate slopes. Both are drawn from within the relevant class rather than sprayed across the map, so the pool is made of genuinely comparable places.

Every site, positive or comparison, then receives the same five features, measured the same way: elevation, local slope, the surrounding ruggedness (the standard deviation of elevation in a 500 m neighbourhood), the mean annual temperature, and the annual precipitation, the last two from WorldClim. We measure every feature the same way: the mean inside a 500 m buffer around each point. That uniformity is a deliberate guard against an artifact we found and discarded during this work. A burst lake's recorded coordinate often sits on the rugged moraine, while a comparison lake's sits on flat water. Measured at the point, a single terrain number then separates the two for a reason that has nothing to do with bursting. A broader set of free streams, including glacier surface velocity, land-surface temperature, and ERA5-Land hydrology [18], was also assembled and offered to the models, but the attribution step kept the compact five above as the susceptibility set. For the triggering axis the features are instead the antecedent weather windows of the first study, the fused CHIRPS--IMERG precipitation and temperature aggregates [8, 12] accumulated over the days and weeks before each date. Each site is evaluated against its own calmer windows rather than against other sites, a case-crossover design borrowed from epidemiology [14]. The design does the heavy lifting on this axis, so it needs explaining. If a site's dangerous window is compared against the same site's quiet windows, then everything permanent about that site, its elevation, its rock, its size, its position in the range, is identical on both sides of the comparison and cancels out. Whatever separation survives has to come from the weather that changed, which is precisely the question the triggering axis asks.

\subsection{The baseline and the five learned models}

The control against which everything else is measured is a deliberately strong but simple model, fitted on the full feature set. Two forms are tried, and the better of them becomes the bar.

The first is a regularized logistic regression. It fits one weight per feature, adds them up, and pushes the total through a curve to get a probability. It can only express the idea that more of a feature means steadily more risk.

The second is gradient-boosted trees [32]. A single decision tree asks a short series of yes-or-no questions about the features and is, on its own, a weak and unstable predictor. Boosting grows hundreds of such trees in sequence, and fits each new tree to the errors the ensemble has made so far, so the ensemble corrects itself step by step. The result captures curved relationships and interactions between features without anyone specifying them in advance. On tabular data of this size it is the standard workhorse, and it is hard to beat. Setting the bar this way separates two things that are easy to confuse. A deeper model that beats a weak baseline may only be showing that the baseline was weak, or that the feature set was good. Requiring the deep model to beat a strong model built on identical inputs and scored on identical folds isolates the contribution of the architecture itself. A learned model that cannot clear that bar does not ship.

Five learned models then face that bar. Each one attacks a specific weakness of this problem. Naming the weakness matters, because a model that fails only teaches us something if its intended job was clear.

\subsubsection{Self-supervised pretraining}

The binding constraint here is labels. We have a few hundred confirmed failures and several thousand lakes we know nothing about. A plain classifier throws the unlabelled majority away.

Self-supervised learning puts them to work. We hide a random quarter of each site's feature values and train an encoder-decoder network to reconstruct what was hidden from what remains. No failure labels enter this stage. To fill in a missing value the network has to learn how the features hang together: that high lakes are cold, that heavy relief accompanies steep slopes, that certain elevation-and-rainfall combinations simply do not occur. That structure is the shape of an ordinary glacial lake. We then discard the decoder, attach a small classifier to the encoder, and fine-tune on the labelled events.

The bet is that most of the learning can happen on unlabelled data, leaving the scarce labels to do only the last and easiest part of the job.

\subsubsection{Label-free anomaly detection}

Our labels are not merely scarce, they are biased. They record where people have looked. Every supervised arm inherits that bias.

This arm avoids labels altogether. We train an autoencoder on the comparison pool alone, squeezing each site through a narrow bottleneck and asking it to reproduce itself. The bottleneck forces the network to keep only what is typical. Then we score every site by reconstruction error, meaning how badly the trained network fails to reproduce it. A site it cannot reproduce is one that does not resemble the ordinary population.

The arm rests on one assumption: that dangerous sites are unusual ones. That assumption is exactly what the experiment tests. Because no failure label is used at any point, this is the only arm that a flawed catalogue cannot corrupt.

\subsubsection{Attention}

The baseline applies one set of feature weights everywhere. Ruggedness counts the same in the Karakoram as in the monsoon Himalaya.

That is probably wrong. An attention layer lets the weights vary by site. For each lake the network computes a small set of multipliers, one per feature, and scales the features by them before classifying. A lake in an arid range can be judged mostly on its relief, and a lake in a wet range mostly on its rainfall, with the model deciding which regime it is in.

The bet is that the drivers of failure change across a region spanning monsoon-fed valleys and cold desert.

\subsubsection{Physics-informed learning}

With a few hundred events and a flexible model, noise can support conclusions that are physically impossible. A network can decide, from a small sample, that steeper slopes are safer.

Slope-stability physics forbids that. Shear demand rises as a slope steepens, so hazard must not fall. We encode the rule directly in the training objective. At each step we differentiate the predicted hazard with respect to the slope feature, which automatic differentiation gives us for free. Where that derivative is negative, meaning hazard is falling as slope rises, we add a penalty proportional to the violation. Where the sign is correct, nothing is charged.

The constraint governs direction, not magnitude. It does not tell the model how much steeper is more dangerous, only that steeper cannot be safer. The bet is that a physically-constrained model generalises better from a small sample than an unconstrained one.

\subsubsection{Positive-unlabeled learning}

This arm addresses the deepest statistical flaw in the setup, and it deserves stating carefully.

Our comparison lakes are not confirmed negatives. A lake with no recorded outburst falls into one of three groups. It may be genuinely stable. It may be dangerous and simply not have failed yet. Or it may have failed at some point without anyone recording it. A standard classifier treats all three as confirmed safe. It is therefore actively taught that dangerous-looking lakes with no entry in the catalogue are safe, which penalises the exact pattern we want it to find.

Positive-unlabeled learning drops that pretence [6]. It treats the unlabelled pool as a mixture of hidden positives and true negatives, in some proportion. The training objective is rewritten accordingly: rather than scoring unlabelled sites as negatives, it estimates the true negative risk by subtracting the positives' expected contribution from the unlabelled set's total. We use the non-negative variant, which clamps that estimate at zero, because on small samples the subtraction can otherwise go negative and the model overfits to the correction itself.

The bet is that correcting for an incomplete catalogue matters more here than any amount of extra model capacity.

\subsubsection{Reading the winner}

Whichever model wins gets opened up rather than reported as a score.

We rank its drivers with Shapley values [31]. The idea comes from cooperative game theory, where the problem is dividing a payout among players who contributed unequally. Here the players are features and the payout is a single prediction. A feature's Shapley value is its average contribution across every possible order in which the features could be added, which makes the attribution fair rather than dependent on where a feature sits in some list. For tree ensembles this can be computed exactly.

The top three features then grow a decision tree. A decision tree splits the data by asking one yes-or-no question at a time, choosing at each step the feature and threshold that best separate failures from non-failures. We cap it at three questions deep, which is shallow enough to print whole and check by eye. Positives are outnumbered, so the classes are weighted to stop the tree from labelling everything as safe.

Attribution alone proves nothing. A feature can take credit for predictions it was never necessary to, if a correlated feature would have served just as well. So the ranking faces a separate test. We delete the top features, refit without them, and see whether skill falls. If it holds up, the model was not really relying on them, and we do not believe the attribution.

\subsection{How skill is measured, and what may be claimed}

Skill is measured by the area under the receiver-operating-characteristic curve, abbreviated ROC. The measure has a plain reading, stated once here and carried through the paper: draw one site that failed and one site that did not, at random, and the ROC is the probability that the model ranks the failed site as the more dangerous of the two. A score of 0.5 is therefore a coin flip, a score of 1.0 is a model that never gets the pair the wrong way round, and a score of 0.76 says the model wins that pairwise comparison about three times in four. The measure earns its place for two reasons. It does not depend on where an alert threshold is set. And it does not move when the ratio of failures to non-failures changes, which matters here: that ratio reflects how many comparison sites we chose to draw, not anything about the mountains.

Each score carries a bootstrap confidence interval, computed from 800 resamples for the susceptibility tests and 1{,}000 for the triggering tests. Every ROC printed in this paper is the median of that bootstrap rather than the single point estimate, which keeps one convention across all of the arms and hazards; the difference between the two is in the third decimal place and never changes a conclusion. The bootstrap answers a question that a single number cannot: how much would the score wobble if we had happened to catch a slightly different sample of events. It works by drawing a new dataset of the same size from the one we have, sampling with replacement so some events appear twice and others not at all, then re-scoring, and repeating until the spread of scores traces out the plausible range. A wide interval is a warning that the estimate rests on few events, and several intervals in this paper are wide for exactly that reason.

Spatial cross-validation governs every score. We group sites into one-degree blocks, roughly a hundred kilometres on a side, and hold out whole blocks together instead of individual sites [34]. The reason is that neighbouring lakes are not independent observations: they share a climate, a rock type, and often a parent glacier, so a model tested on a lake whose neighbour it trained on can score well by recognising the neighbourhood rather than by understanding the hazard. Holding out whole blocks is the equivalent of testing a student on a chapter they never studied instead of on the practice problems they memorised. It produces a lower and less flattering number than ordinary random splitting, and it is the only number reported here.

Three rules then govern what may be claimed. \label{para:rules} A learned model beats the simple control only when its confidence interval clears the control's, so a win of a few hundredths inside the noise counts for nothing. A signal counts as real only when the control itself clears chance on held-out blocks. And when a separation looks suspiciously large, we check it against the raw feature distributions and throw it out if one feature is carrying the result for a measurement reason rather than a physical one. That check uses a standardized separation: the gap between the two groups' averages, expressed in standard deviations, so features in metres and features in millimetres sit on one scale. Anything above two standard deviations on a single feature is a red flag, not a discovery. Where a result is a null, it is reported as a null. A provenance ledger records every input and every derived table. Every headline number here is checked against its committed source in an accompanying notebook, which fails loudly if the paper and the data ever disagree.

\subsection{How the key steps are implemented}

The design above is only as good as its implementation, and four pieces of that implementation carry most of the weight. Each is short enough to show in full, and showing it is the point: a reader who doubts a result should be able to see the operation that produced it rather than trust a description of it. All of it is ordinary Python built on \texttt{scikit-learn}, run on a laptop; nothing here needs a cluster.

\subsubsection{Matching events to nearby lakes}

The matched control is the single most consequential operation in the paper, because it is what separates a real terrain signal from the catalogue-geography artifact. The rule is simple: keep a comparison lake only if it lies within 50 km of at least one lake that actually burst, and require both groups to clear the same elevation floor so the comparison is not quietly one of high lakes against low ones.

\begin{lstlisting}
# keep only comparison sites within `matched_km` of some real failure
posll = sub[sub.story == poslabel][["lat", "lon"]].values
cand  = sub[sub.story == candlabel]
dmin  = np.array([haversine_km(r.lat, r.lon, posll[:, 0], posll[:, 1]).min()
                  for r in cand.itertuples()])
sub = pd.concat([sub[sub.story == poslabel],
                 sub.loc[cand.index[dmin <= matched_km]]])
\end{lstlisting}

Here \texttt{dmin} is each candidate's distance to the nearest failure. The filter keeps only candidates sitting in the same neighbourhoods as the events. Figure~\ref{fig:geo} is what happens when the identical scoring code runs with and without those lines. The score changes because the comparison set changed, never because the model did.

\subsubsection{Scoring without leaking geography}

Every score in the paper comes from one function, so that no arm can quietly get an easier evaluation than another. It groups sites into one-degree blocks, holds whole blocks out together, accumulates predictions out of fold, and builds the confidence interval by resampling them.

\begin{lstlisting}
p = np.zeros(len(y))
for tr, te in GroupKFold(5).split(X, y, grp):          # grp = 1-degree block id
    p[te] = GradientBoostingClassifier(random_state=0).fit(X[tr], y[tr]) \
              .predict_proba(X[te])[:, 1]
boots = []
rng = np.random.RandomState(0)
for _ in range(800):                                    # bootstrap the ROC
    ix = rng.randint(0, len(y), len(y))
    if len(np.unique(y[ix])) > 1:
        boots.append(roc_auc_score(y[ix], p[ix]))
roc, lo, hi = np.median(boots), *np.quantile(boots, [.025, .975])
\end{lstlisting}

Two details are deliberate. The fold split is keyed on the block, not the site, which is what stops a lake being scored by a model that trained on its neighbour. And the seeds are fixed, so the numbers in this paper are reproducible to the digit rather than merely to the confidence interval.

\subsubsection{The physics penalty}

The physics-informed network differs from an ordinary network by one term in its loss. Slope-stability physics says hazard must not fall as a slope steepens. So the loss charges the model whenever the gradient of its predicted hazard with respect to slope turns negative.

\begin{lstlisting}
def physics_penalty(model, x):
    x = x.clone().requires_grad_(True)
    hazard = torch.softmax(model(x), 1)[:, 1].sum()
    grad = torch.autograd.grad(hazard, x, create_graph=True)[0]
    return 0.5 * torch.relu(-grad[:, SLOPE_IDX]).mean()   # only wrong-signed slopes cost
\end{lstlisting}

The \texttt{relu} makes the term a guardrail rather than a target. A correctly-signed relationship costs nothing. Only the physically impossible direction is charged. The intent is to stop a model trained on few events from learning a relationship no geomorphologist would accept. Honesty requires adding that the guardrail bought no extra skill here.

\subsubsection{Confirming attribution by deletion}

Attribution is a claim about what a model uses, and a claim of that kind should be tested rather than displayed. The test is to delete the features that Shapley attribution ranked highest, refit on what remains, and see whether skill actually falls.

\begin{lstlisting}
top_idx  = np.argsort(np.abs(shap_values).mean(0))[::-1][:3]   # 3 strongest features
rest_idx = [i for i in range(X.shape[1]) if i not in top_idx]
roc_all     = spatial_cv_roc(X, y, grp)                # every feature
roc_ablated = spatial_cv_roc(X[:, rest_idx], y, grp)   # top three removed
confirms_shap = (roc_all - roc_ablated) > 0.02
\end{lstlisting}

A model whose skill survives the deletion of its supposedly most important features was not really using them, and the attribution should not be believed. The burst models fail that deletion sharply. That is the basis for the claim in R5. The landslide model does not fail it, so the same claim is withheld there.
\section{Results}

\begin{figure}[H]
\centering
\includegraphics[width=0.86\linewidth]{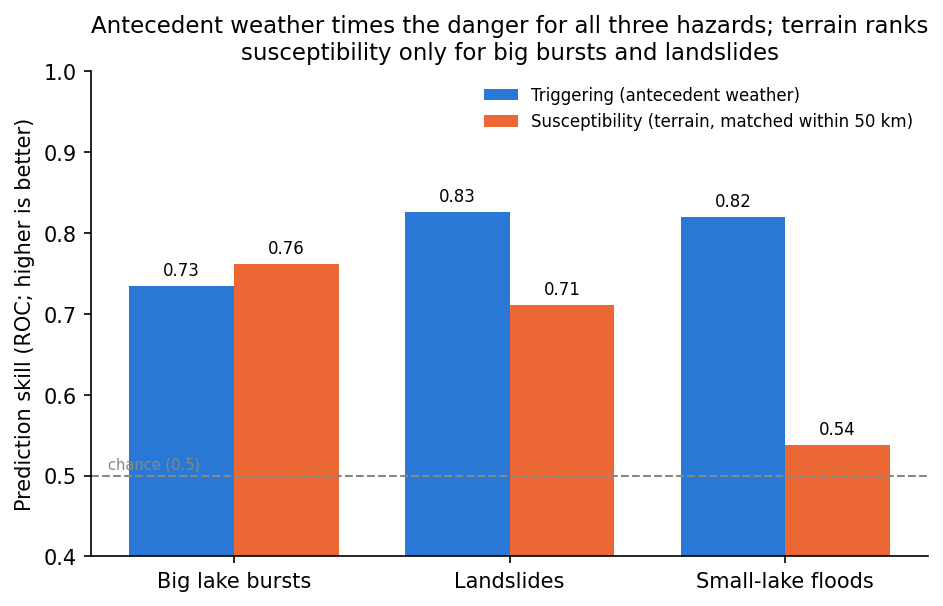}
\caption{Weather predicts the dangerous windows across all three hazards; terrain ranks susceptibility only for the big bursts and landslides, and not for the small floods.}
\label{fig:axes}
\end{figure}

The results follow the two axes. Table 1 and Figure~\ref{fig:axes} are the scoreboard: for each hazard, how well antecedent weather times the trigger and how well terrain ranks susceptibility. The subsections take the findings in turn.

Every result below is a single number between 0.5 and 1.0. Fixing what those numbers mean, before any of them arrive, makes the rest easier to read. Take two sites, one that failed and one that did not. Ask the model which of the pair is more dangerous. A score of 0.5 means it is guessing; it gets the pair right half the time, exactly as a coin would. A score of 1.0 means it is never wrong. A score of 0.76 means it wins about three of every four such comparisons, and loses the fourth. The field calls these areas under the receiver-operating-characteristic curve. Nothing below depends on knowing that. The pairwise reading is the whole of it.

Two habits of reading follow. First, a score can be useful well short of 1.0. A list that puts dangerous lakes near the top three times in four beats inspecting lakes in arbitrary order, and it still beats it despite being wrong about a quarter of the time. Second, a score is only as good as the comparison behind it. That is why most of what follows concerns who the failed sites were measured against, not the models.

\begin{table}[H]
\centering
\caption{The two axes of prediction, by hazard. WHEN is the weather-trigger skill on held-out windows; WHICH-site is the terrain skill after each site is compared only to comparable sites within 50 km. All values are ROC under spatial cross-validation; 0.5 is chance.}
\begin{tabularx}{\linewidth}{Xccc}
\toprule
\textbf{Hazard} & \textbf{WHEN (weather)} & \textbf{WHICH-site (terrain)} & \textbf{Which-site verdict} \\
\midrule
Big lake bursts (moraine / ice) & 0.73 & 0.76 & real, modest \\
Landslides & 0.83 & 0.71 & real, modest \\
Small-lake floods & 0.82 & 0.54 & no signal \\
\bottomrule
\end{tabularx}
\end{table}

\subsection{R1. Antecedent weather identifies the dangerous windows}

Triggering is the stronger of the two axes. A simple model on the antecedent-weather windows separates a dangerous window from a calm one at ROC 0.73 for the big bursts, 0.83 for the landslides, and 0.82 for the small floods. The ordering is physically sensible: the small floods and the landslides are the most directly rainfall- and melt-driven, while the big moraine bursts include the avalanche-triggered failures that no weather signal can see. The gap between 0.73 and 0.83 is therefore not a modelling shortfall to be engineered away. It is the trigger ceiling the literature describes [28, 9]. One family of triggers is partly forecastable from the weather. The other, a rock or ice avalanche falling into the lake on a clear day, is not. No weather model can see a trigger that leaves no weather signature, and the big-burst population contains more of those triggers than the other two hazards do. The drivers are antecedent precipitation and temperature, as before. This axis is regional and seasonal, not lake-specific: it says the monsoon has primed the region, not which lake will go.

\subsection{R2. Terrain ranks susceptibility, once a geographic artifact is removed}

\begin{figure}[H]
\centering
\includegraphics[width=0.92\linewidth]{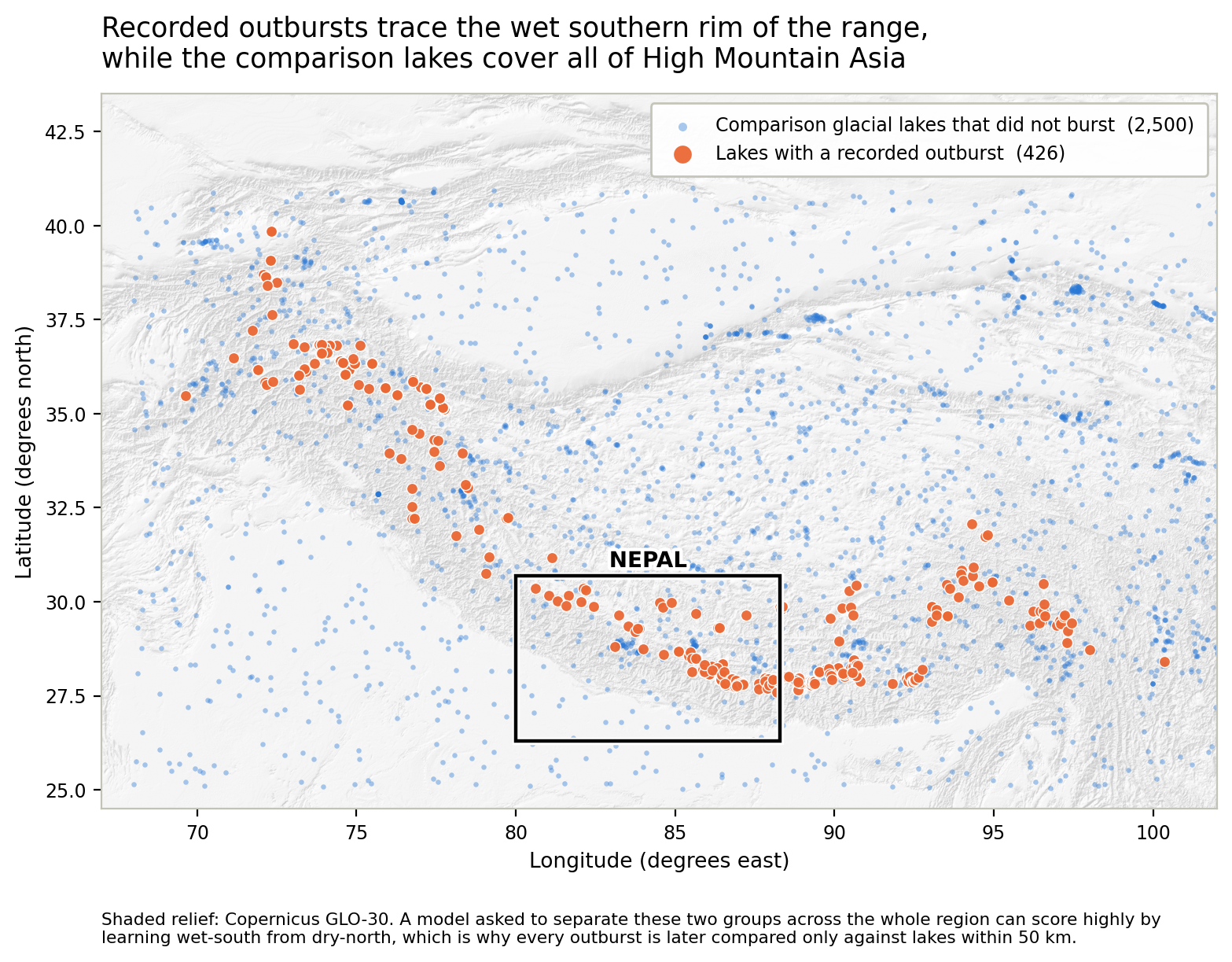}
\caption{Where the events are, and where the comparison lakes are. Recorded bursts (red) follow the wet Himalayan and Karakoram arc; the comparison lakes (grey) span the whole region, including the arid Tibetan interior. A model asked to separate the two can succeed by learning geography instead of hazard, which is why every burst is subsequently compared only against lakes within 50 km.}
\label{fig:map}
\end{figure}

Asked naively across all of High Mountain Asia, terrain appears to rank susceptibility almost perfectly: ROC 0.92 for the big bursts, 0.91 for the small floods, 0.98 for the landslides. Those scores are an artifact, and the artifact is geographic. The catalogued events cluster in the wet, monsoon-fed ranges where floods are both frequent and observed, while the comparison pool spans the whole region, including the cold, dry Tibetan interior where lakes are plentiful and recorded outbursts are scarce. A model can score 0.9 simply by learning warm-and-wet from cold-and-dry, which is a map of where people look, not of which lake bursts. The mismatch is visible on the map (Figure~\ref{fig:map}). The recorded bursts trace a narrow arc along the wet southern rim of the range. The comparison lakes spread across the whole of High Mountain Asia, including the cold, dry interior of the Tibetan Plateau, where lakes are common and catalogued outbursts are not.

\begin{figure}[H]
\centering
\includegraphics[width=0.86\linewidth]{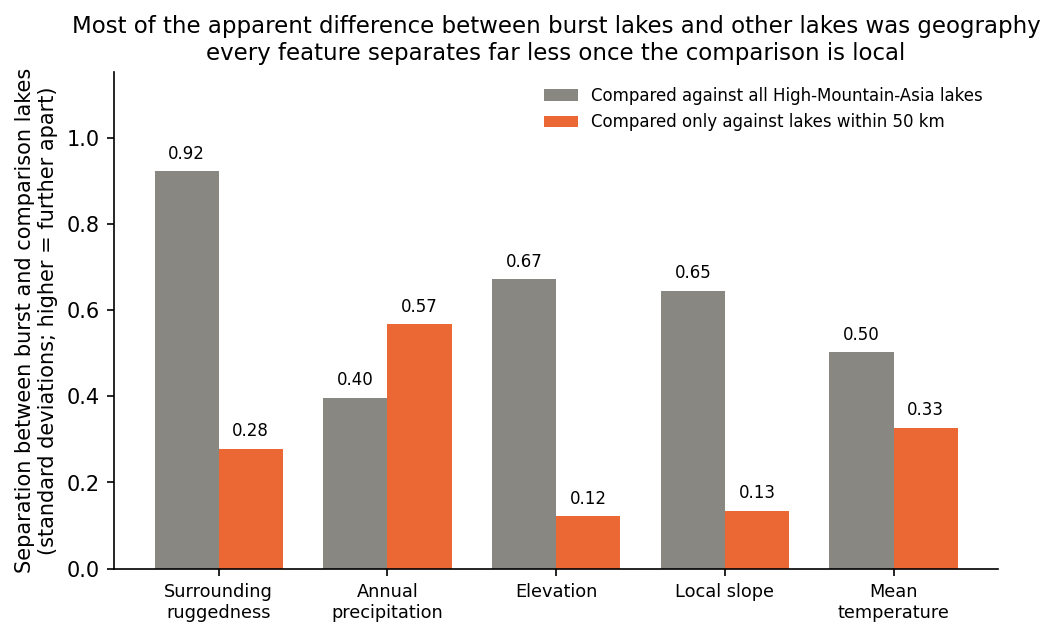}
\caption{How far apart burst lakes and comparison lakes really are, feature by feature, before and after the comparison is restricted to lakes within 50 km. Separation is measured in standard deviations. Most of the apparent difference was a difference of geography rather than of hazard.}
\label{fig:sep}
\end{figure}

\begin{figure}[H]
\centering
\includegraphics[width=0.86\linewidth]{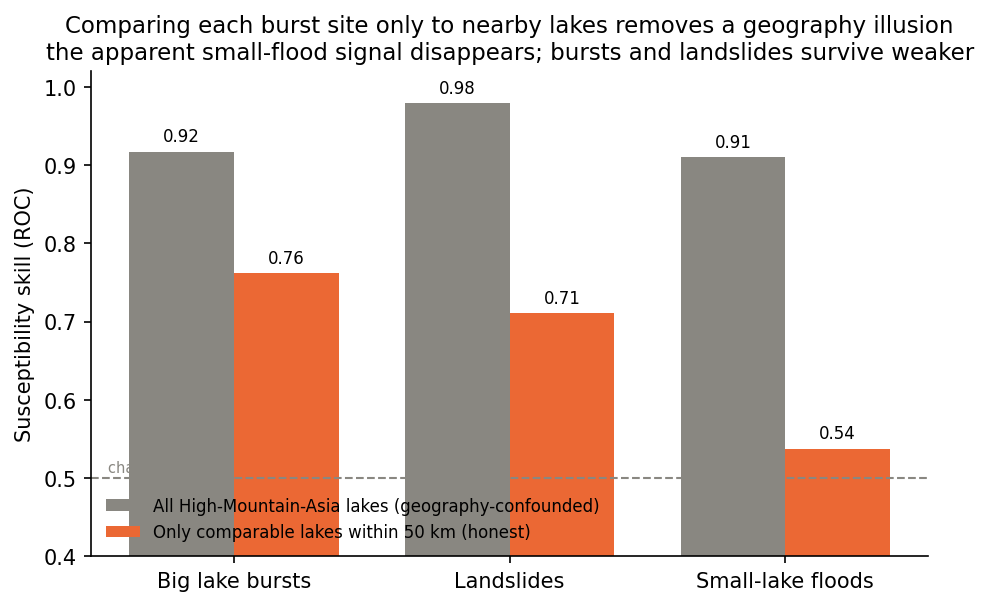}
\caption{Comparing each event only against comparable sites within 50 km removes the geographic artifact. The apparent small-flood signal disappears entirely; the burst and landslide signals survive at a weaker, honest level.}
\label{fig:geo}
\end{figure}

The control that settles it is to compare each event only against comparable sites within 50 km, in the same climate and at the same elevation. Under that control the inflated scores collapse to their honest level (Figure~\ref{fig:geo}): 0.76 for the big bursts, 0.71 for the landslides, and 0.54, no better than chance, for the small floods. State the size of that fall plainly. At 0.92 the model won better than nine of every ten paired comparisons. At 0.76 it wins about three in four. At 0.54 it is barely beating a coin. What disappeared was not a refinement at the margins. It was most of the apparent performance, and the model had been earning it by recognising which part of the map it was looking at. The collapse is largest exactly where the confound was largest. Every feature separates the two groups far less once the comparison is local (Figure~\ref{fig:sep}): the standardized separation on the strongest single feature falls from more than two standard deviations to well under one, and no feature retains a separation large enough to carry a result on its own. The pattern is what a removed confound looks like. Had the original separation been physical, restricting the comparison to nearby lakes of the same climate would have left it largely intact, because the physics of a moraine dam does not change across fifty kilometres.

 The point generalises beyond this study. Published susceptibility models rank glacial lakes at around 0.83 [13], but usually against an unmatched background of lakes that did not fail. Part of that skill, like ours before the control, may reflect where events are catalogued rather than which lake is primed.

\subsection{R3. The burst signal is real inside a single region, and depends on the failure mechanism}

\begin{figure}[H]
\centering
\includegraphics[width=0.82\linewidth]{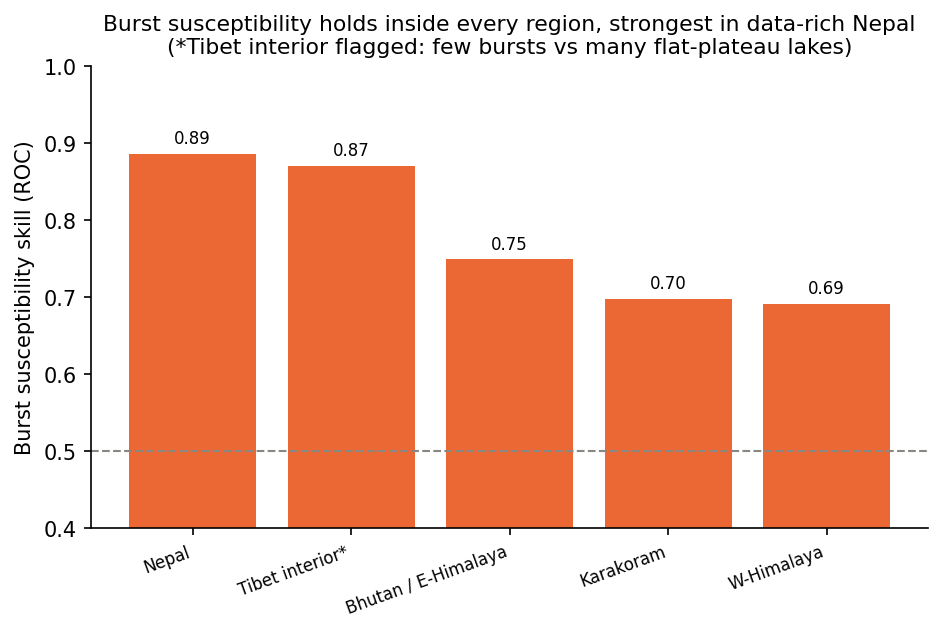}
\caption{The burst susceptibility signal holds inside every region, strongest in data-rich Nepal, so it is a real within-region signal rather than a between-region artifact. The Tibetan-interior bar is flagged (few events against many flat-plateau lakes).}
\label{fig:region}
\end{figure}

A skeptic can still ask whether the surviving 0.76 is itself a milder version of the same between-region effect, since a fifty-kilometre neighbourhood is local but not tiny. The answer is no, and the test that settles it removes between-region contrast entirely. Run the model inside one region at a time, so every comparison happens among lakes of the same range. The burst signal holds in every region tested. It is strongest where the data are richest: 0.89 within Nepal alone, then 0.75 in Bhutan and the eastern Himalaya, 0.70 in the Karakoram, and 0.69 in the western Himalaya (Figure~\ref{fig:region}). That the within-Nepal figure exceeds the pooled 0.76 shows the pooled number was diluted by the harder western ranges, not inflated by mixing them.

\begin{figure}[H]
\centering
\includegraphics[width=0.78\linewidth]{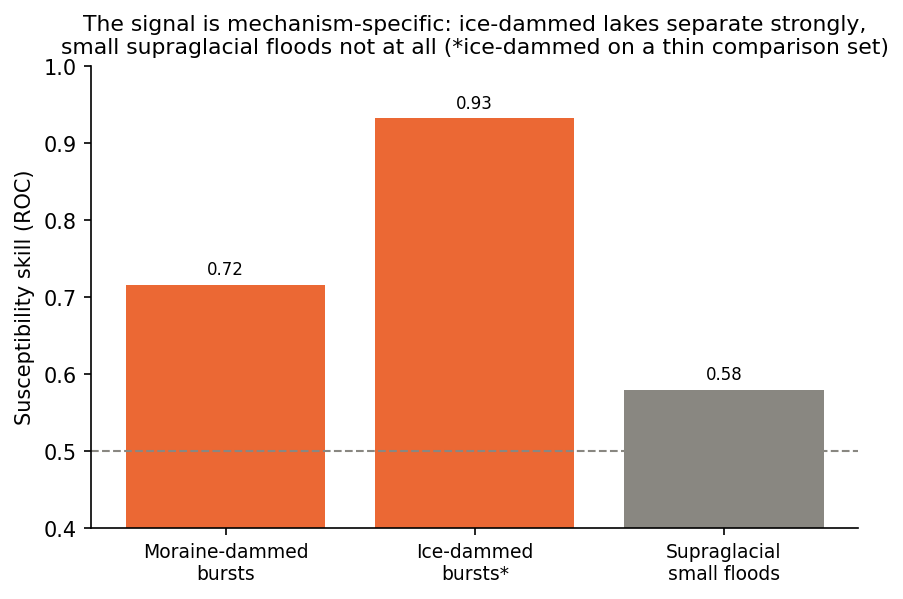}
\caption{The signal depends on the failure mechanism: ice-dammed lakes separate strongly, moraine-dammed bursts modestly, and small supraglacial floods not at all. The ice-dammed estimate rests on a thin comparison set.}
\label{fig:type}
\end{figure}

The signal is also mechanism-specific (Figure~\ref{fig:type}). Split by dam type, ice-dammed lakes are the most separable, at 0.93, because they sit in the distinctive terrain of a glacier snout; the moraine-dammed bursts, the catastrophic ones, separate modestly at 0.72; and the supraglacial small floods do not separate at all, at 0.58, even compared only against their own kind. The ice-dammed result is a regional one, since Nepal's dangerous lakes are almost entirely moraine-dammed [21]; the moraine figure therefore carries the Nepal watchlist, and we report the ice-dammed separability as a wider-region observation. The small-flood null is therefore robust: it holds within region and within mechanism. One caution is carried forward honestly: the Tibetan-interior burst estimate, though high, rests on few events against many flat-plateau lakes and leans on a single ruggedness gradient, and is flagged rather than trusted.

\subsection{R4. Where the deep models led, and why we still do not declare a winner}

\begin{figure}[H]
\centering
\includegraphics[width=0.86\linewidth]{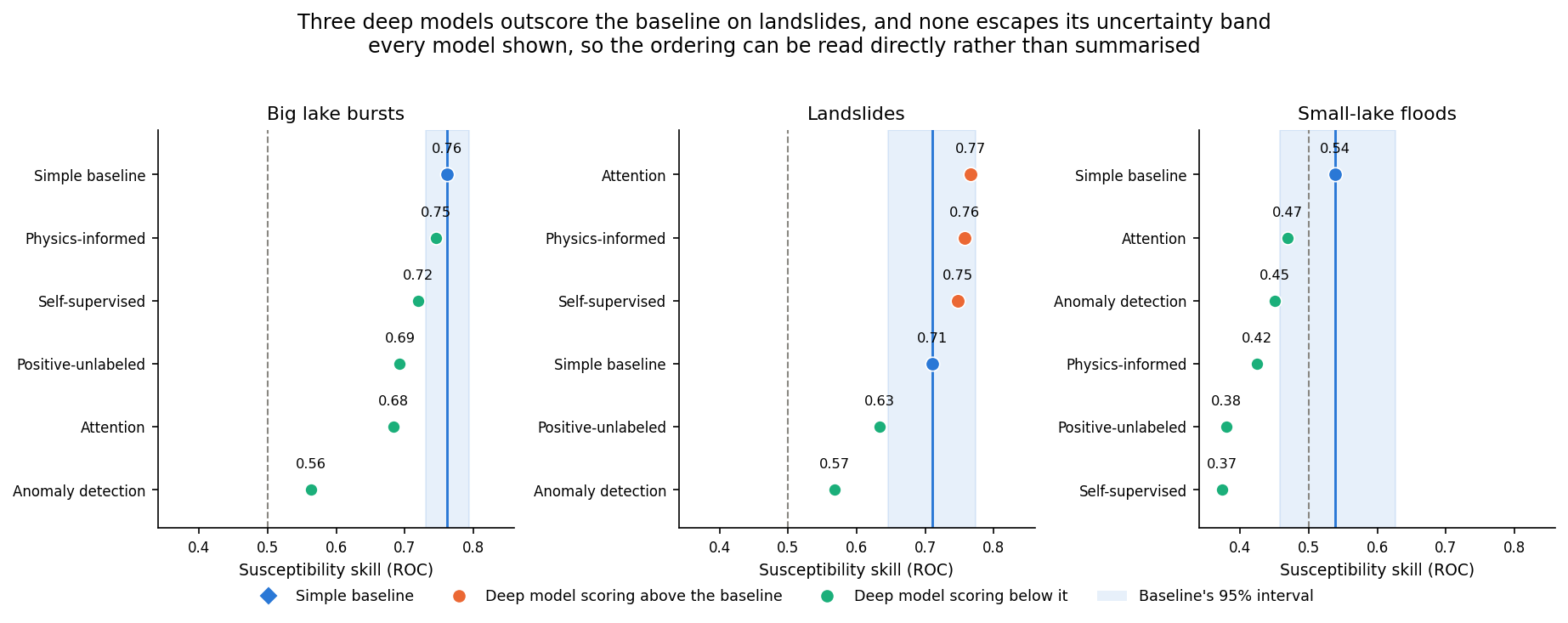}
\caption{Even the best of five deep models does not clearly beat the simple gradient-boosted baseline in any hazard; no deep model clears the baseline's confidence interval.}
\label{fig:deep}
\end{figure}

Table 2 gives every model on every hazard, and Figure~\ref{fig:deep} plots the same numbers against the baseline's uncertainty band. We show all of them rather than a best-of summary, because the ordering carries information that a summary would hide.

\begin{table}[H]
\centering
\caption{Susceptibility skill for all six models on all three hazards, matched within 50 km. Bold marks a deep model scoring above the simple baseline. All values are bootstrap-median ROC under spatial cross-validation.}
\begin{tabularx}{\linewidth}{Xccc}
\toprule
\textbf{Model} & \textbf{Big lake bursts} & \textbf{Landslides} & \textbf{Small-lake floods} \\
\midrule
Simple baseline (boosted trees) & 0.76 & 0.71 & 0.54 \\
Self-supervised pretraining & 0.72 & \textbf{0.75} & 0.37 \\
Anomaly detection & 0.56 & 0.57 & 0.45 \\
Attention & 0.68 & \textbf{0.77} & 0.47 \\
Physics-informed & 0.75 & \textbf{0.76} & 0.42 \\
Positive-unlabeled & 0.69 & 0.63 & 0.38 \\
\bottomrule
\end{tabularx}
\end{table}

The landslide column is the interesting one. Three of the five deep models score above the baseline there: attention at 0.77, physics-informed at 0.76, and self-supervised pretraining at 0.75, against the baseline's 0.71. The margins run from three to six hundredths. More telling than any single margin is the pattern. Three architectures built on different principles all landed on the same side, and the two that did not are the two whose assumptions the data contradicts elsewhere, since anomaly detection assumes danger equals unusualness and the positive-unlabeled correction matters most where catalogues are thin.

We still do not call it a win, and the reason is the sample rather than the models. The baseline's interval on landslides runs from 0.65 to 0.77, and all three leading models fall inside it. An interval that wide means the evidence is consistent with the deep models being modestly better, no different, or slightly worse. Declaring a winner from a margin that sits inside the noise is how a field accumulates improvements that later fail to replicate.

The honest reading is therefore neither that depth failed nor that depth won. On the two lake hazards the baseline is ahead and the deep models add nothing, which is a real result at these sample sizes. On landslides there is a consistent, repeated hint that depth helps, which this dataset cannot resolve. Landslides also happen to be the hazard with by far the most catalogued events, so it is the one where the question can actually be settled. We treat it as the most promising open lead in the paper rather than as a null, and say so in the next steps.

One further case deserves flagging rather than counting. In the Tibetan interior a deep model does clear the baseline's interval outright. That is also the least trustworthy segment in the study, with 38 events against 754 flat-plateau lakes and a strong ruggedness gradient, so we read it as the networks exploiting the artifact described in R3 rather than as evidence for depth.

\subsection{R5. What the burst model actually uses}

\begin{figure}[H]
\centering
\includegraphics[width=0.80\linewidth]{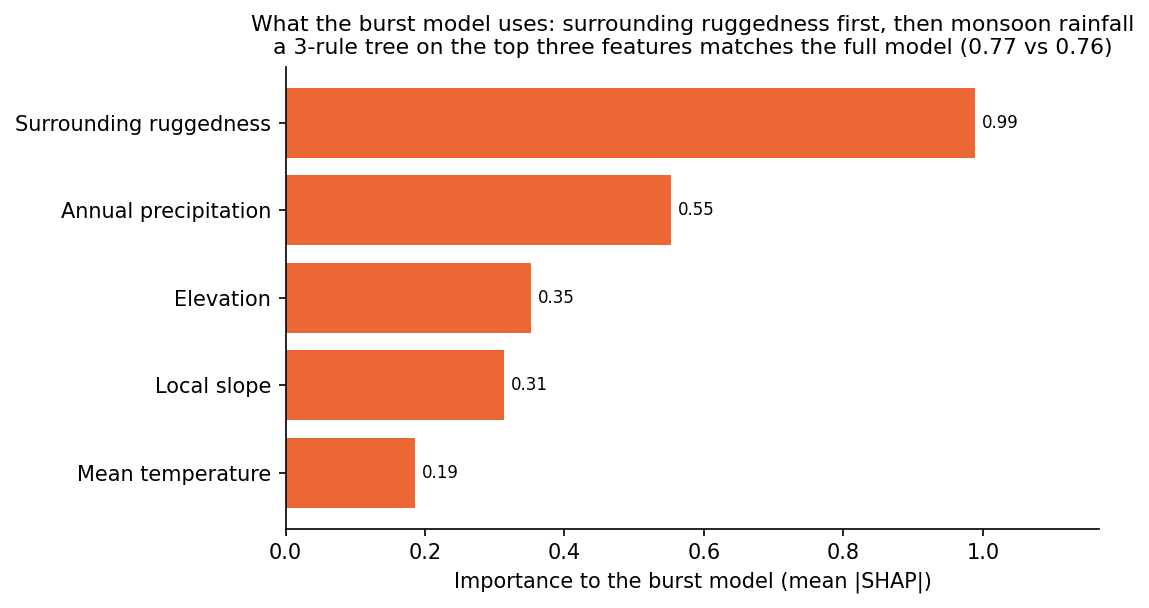}
\caption{What the burst model uses: surrounding ruggedness first, then monsoon rainfall. A three-rule decision tree on the top three features matches the full model, and an ablation confirms the ranking.}
\label{fig:shap}
\end{figure}

\begin{figure}[H]
\centering
\includegraphics[width=\linewidth]{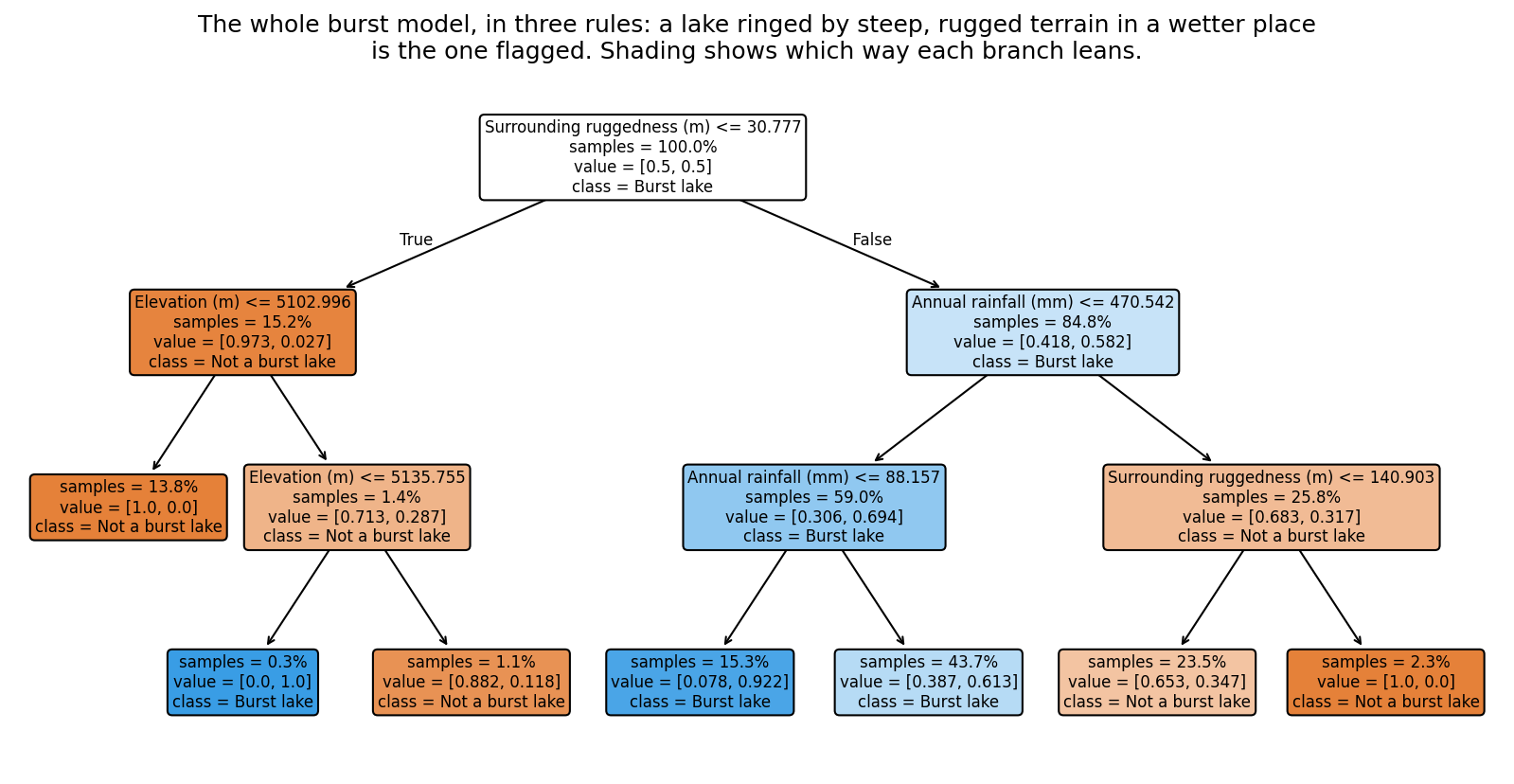}
\caption{The entire burst model written out as three questions. Each box gives the rule, the share of sites reaching it, and which way that group leans; blue leans toward burst lakes and orange away from them. A lake in flat surroundings is dismissed at the first question, and among rugged settings the wetter ones are flagged.}
\label{fig:tree}
\end{figure}

Because the winning model is simple, it can be read. The Shapley attribution ranks surrounding ruggedness first, then annual precipitation, then elevation (Figure~\ref{fig:shap}), a physically sensible set: a lake ringed by steep, high relief in a wet place is the one primed to fail. The reading survives its own test, in two independent ways. First, a decision tree only three questions deep, built on those three features alone, reproduces the full model for the pooled bursts, scoring 0.77 against the full model's 0.76. The tree is small enough to print in full (Figure~\ref{fig:tree}), and reading it recovers a physical story rather than a statistical curiosity. The first split sets aside lakes in flat surroundings, which are almost never burst sites. Among the rugged ones, the wetter settings are what the model flags. A reader who disagrees with the model can therefore point at the exact rule they disagree with, which is not possible with any of the five deep alternatives.

Second, the ranking is confirmed by deletion. Removing the top three features and refitting costs a tenth of a point of skill on the pooled bursts, and nearly two tenths within Nepal. The deletion test matters because attribution alone can be misleading: a feature can be handed credit for predictions it was never necessary to, if a correlated feature would have served just as well. A skill drop of that size shows these three are load-bearing rather than decorative. The landslide model is the exception, and it should be said plainly. Its skill spreads redundantly across correlated climate and terrain features. No small set is uniquely responsible, and dropping the top three barely moves it. Its signal is real but diffuse, closer to a broad climate-and-terrain gradient than to a sharp rule.

\subsection{R6. Deformation marks hazard but does not rank susceptibility}

For completeness the interferometric deformation signal, the precursor the feasibility study measured directly and the one the recent literature ties most closely to imminent dam failure [3, 29], was put to the same susceptibility test against a control set of dangerous-but-unfailed lakes. It did not separate them: burst lakes and flagged-but-stable lakes deform alike, and a within-lake test found no acceleration before a lake's own failure beyond its own quiet periods. The distinction being drawn here is fine and easy to lose, so here it is twice over. Moraine dams that are slowly sagging are indeed more dangerous than dams that are not, and the feasibility study measured that sagging directly. What the present test asks is a harder question: among lakes that have already been flagged as dangerous, does the amount of sagging tell you which one will actually go. The answer is no. The dangerous-but-stable lakes are sagging too. A signal can be a genuine mark of hazard and still be useless for ranking susceptibility. A persistent cough is a real sign of illness, and a poor way to pick which patient in a waiting room will deteriorate first. Deformation is therefore a necessary-not-sufficient hazard marker, to be fused with the weather trigger rather than used alone to rank lakes, which is consistent with its role as a measure of a dam that is slowly degrading rather than of one that is about to be triggered. Sharpening it would need the full small-baseline interferometric inversion [2] the feasibility study specified, not the light chain used here.

\subsection{R7. A Nepal watchlist}

\begin{figure}[H]
\centering
\includegraphics[width=\linewidth]{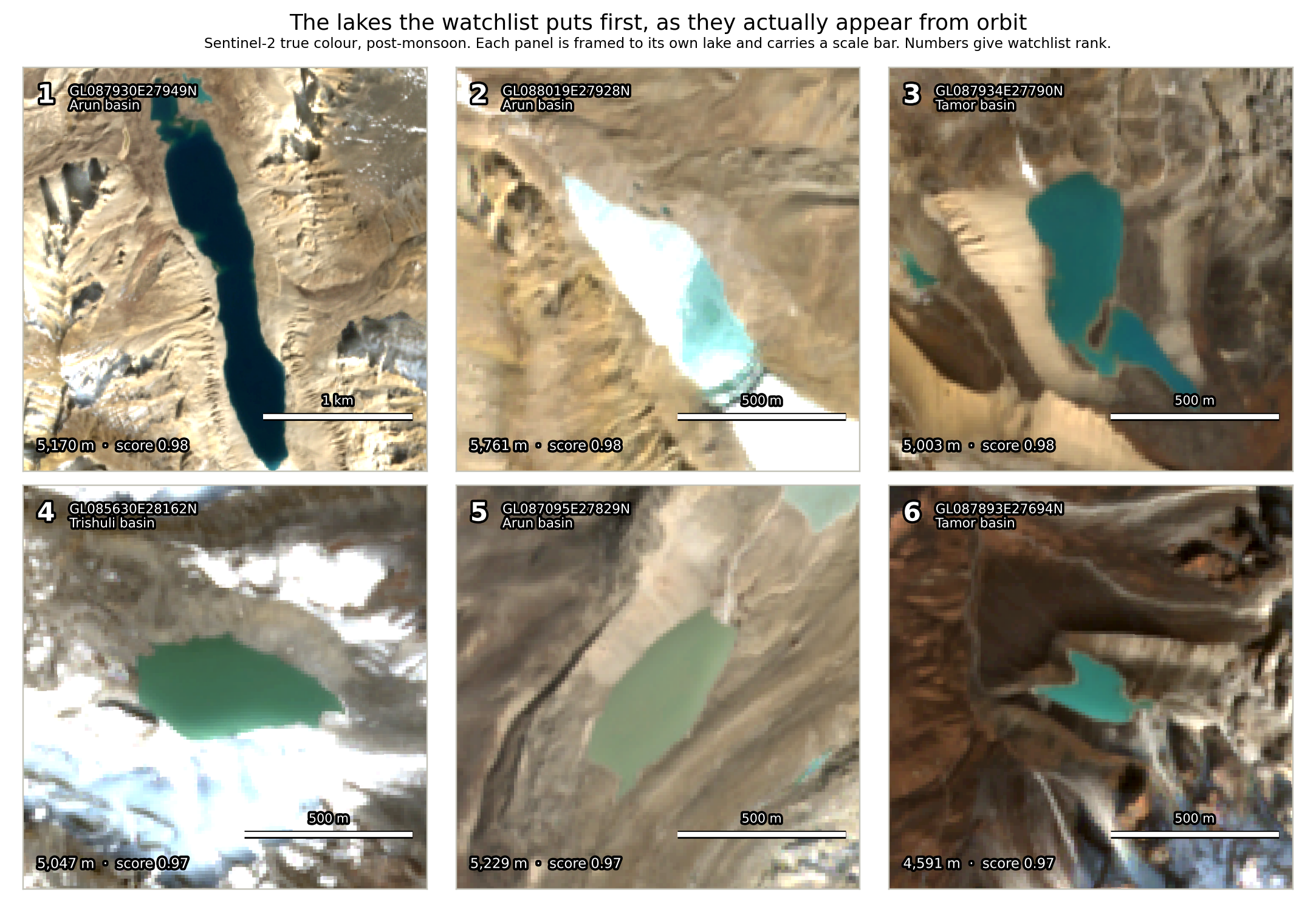}
\caption{The six highest-ranked lakes on the Nepal watchlist, in Sentinel-2 true colour. Each panel is framed to its own lake and carries a scale bar, so the lakes can be compared despite differing greatly in size. The imagery is shown so that the ranking can be inspected rather than taken on trust; it is not itself evidence of instability.}
\label{fig:imagery}
\end{figure}

\begin{figure}[H]
\centering
\includegraphics[width=0.80\linewidth]{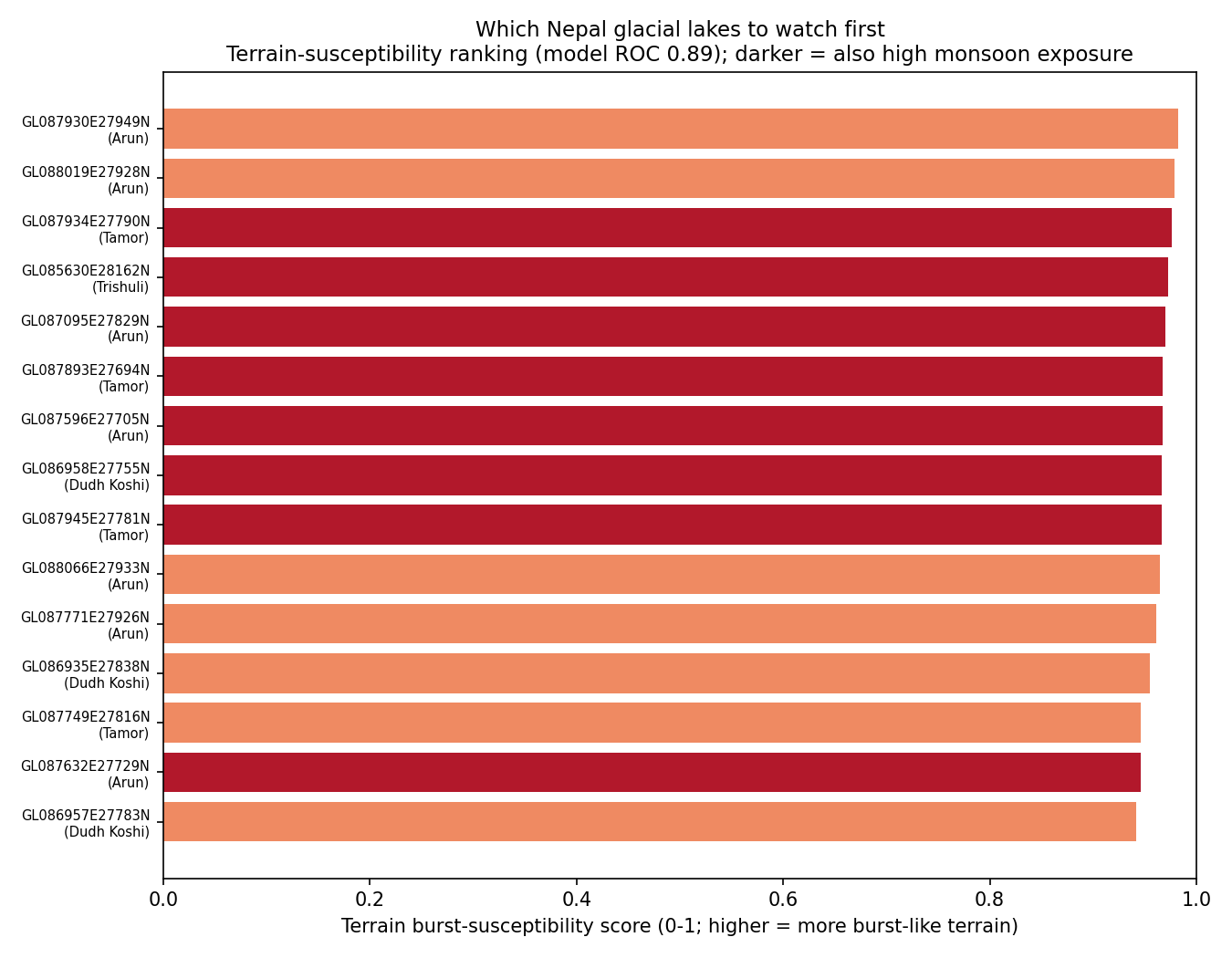}
\caption{The Nepal watchlist: named potentially-dangerous lakes ranked by terrain burst-susceptibility (model ROC 0.89), with darker bars marking lakes that also carry high monsoon exposure. A prioritisation aid, not a prediction.}
\label{fig:watch}
\end{figure}

The parts that survived the test combine into the deliverable the whole effort was for: a ranked watchlist for Nepal. We take the within-Nepal burst model, the strongest susceptibility result at 0.89, and run it over the 47 named lakes in ICIMOD's inventory of potentially-dangerous glacial lakes. Overlaying its terrain score with each lake's monsoon exposure gives a combined susceptibility-and-trigger ranking (Figure~\ref{fig:watch}). Its ordering agrees broadly with ICIMOD's own hazard rank. That agreement is a validation, not a coincidence. The lakes it puts on top are the high, steep-sided ones of the Arun, Tamor, and Dudh Koshi basins. The list is offered for what it is: a modest-skill prioritisation to help decide where to look first, not a prediction that any particular lake will burst, and a starting point for the institutions that hold the mandate. Be concrete about what such a list is good for, because a ranking invites more confidence than it has earned. An agency can afford to send a survey team to a handful of lakes in a season. Those lakes have to be chosen somehow. A ranking that gets nine of every ten paired comparisons right within Nepal is a better basis for that choice than lake size, proximity to a road, or the order in which lakes were last visited. It is not a basis for telling a valley that its lake is about to fail. The list says where to look first. The looking is still what establishes whether a lake is dangerous. A ranking is easier to argue with when you can see the places in it. Figure~\ref{fig:imagery} shows the top six as they appear from orbit. They share the setting the model learned to look for: meltwater ponded against a moraine at the foot of a glacier, ringed by steep rock and ice, high enough that several stay part-frozen through the post-monsoon window.

\section{Discussion}

The prediction test sorts the two questions cleanly, and the sorting is the contribution. Triggering is the more answerable of the two: free satellite weather marks the dangerous window for every hazard at a skill of roughly 0.8, bounded only by the avalanche-triggered bursts that arrive on a clear day. Susceptibility is answerable only in part, and only for the catastrophic hazards. Terrain ranks a primed lake or slope at a modest 0.7 to 0.76 across the region, rising to 0.89 within Nepal. It cannot rank the small floods at all. A risk layer built on free data should therefore lean on triggering for its daily signal and treat susceptibility as a slow prior over sites, which is precisely the risk-equals-probability-times-impact architecture the first study arrived at, now measured on both axes rather than argued.

Two results are worth stating plainly because they run against the grain of the field's enthusiasm for larger models. First, a near-perfect susceptibility score is available for the asking and is almost entirely an artifact of where events are catalogued; only a matched control exposes the honest, weaker signal underneath, and any susceptibility study that ranks failed sites against an un-matched background pool should be read with that in mind. Second, no deep model decisively beat a gradient-boosted tree on five features, across three hazards, two axes and five modern architectures. On landslides three of them did score above it, consistently enough to be worth chasing, but never by more than the noise allows. And on the hazard where the baseline wins outright, the winner reduces to three rules a hydrologist can read. The scarce quantity here is labeled failures and clean features, not model capacity, and the honest response is a transparent model with stated uncertainty rather than a deeper one with borrowed confidence.

The limits are the same ones the first study named, and they bound these numbers too. The event catalogues are spatially uneven, which is the very confound the matched control had to remove and which still shapes what ``comparable'' can mean. The comparison pool is unlabeled rather than known-safe, so a small and accepted fraction of its lakes may be dangerous-but-not-yet-failed. The susceptibility features are terrain and climate only; the quantity the susceptibility literature ranks highest, lake volume [13], is not observable from a surface image and must be surveyed or scaled from bathymetry, and the deformation channel that would most directly measure a degrading dam is a hazard marker rather than a ranker. Each is a limit of free data and light processing. Each would ease with the partnership data the first study specified. Local weather and lake telemetry would lift the triggering ceiling. Bathymetry would add the volume axis. Full interferometric processing would sharpen the deformation channel.

Within those limits the picture is coherent across the whole effort. The lake floods that the area detector could not label are the same small floods that terrain cannot rank, and both nulls point the same way: the small, frequent events are governed by triggers and englacial plumbing that free static data does not see. The catastrophic bursts are different. They leave a terrain fingerprint modest enough to prioritise on, and a weather signature timely enough to watch. A free-data screening layer can honestly be built on those.
\section{Conclusions and Next Steps}

Turning the feasibility assessment into a prediction test gives a clear and honest answer. From free satellite data, we can time the dangerous windows at a Himalayan site with useful skill, for all three hazards. We can rank which lake is susceptible only modestly, and only for the catastrophic bursts and the landslides. Within Nepal that rises to 0.89. For the small, frequent floods we cannot say at all. The skill that exists belongs to a simple, transparent model, and the useful product is a ranked Nepal watchlist offered as a prioritisation for the institutions that already hold the mandate.

\subsection{What this work contributes}

Six things come out of it, and they are worth listing plainly.

The first is a control that changes how susceptibility scores should be read. Published models rank glacial lakes at around 0.83 [13], but generally against a background of lakes drawn from wherever they happen to be. We show what that costs. On our own data an unmatched background returns 0.92, and matching each burst to lakes within 50 km takes the same model, the same features and the same folds down to 0.76. The gap is not a quirk of our pipeline. It is a property of comparing catalogued failures against an uncatalogued world, and any study using an unmatched background inherits it. The fix is cheap to apply and we give the code.

The second is a working separation of susceptibility from triggering. The two are usually blended into a single hazard score. Measuring them apart, on one dataset and one protocol, shows they behave very differently: triggering is the more forecastable of the two, susceptibility the weaker, and the two fail for different reasons.

The third is a three-hazard benchmark. Lake bursts, landslides and small glacial floods are normally studied by separate communities. Running all three through identical folds, an identical baseline and an identical gate makes their skills directly comparable, and the comparison is itself informative: the small floods fail everywhere the others succeed.

The fourth is a susceptibility model an institution can audit. It is three rules on three features, printed in full in this paper. An engineer who disagrees with it can point at the rule they dispute. That is not true of any of the five deep alternatives, and it matters for a model meant to inform public decisions.

The fifth is a ranked, named Nepal watchlist built from data anyone can download, with its skill stated rather than implied, and with its agreement against ICIMOD's independent hazard ranking reported.

The sixth is a set of honest nulls. Small glacial floods do not separate from free static data, in any region and by any mechanism. Deformation marks hazard without ranking it. Five modern architectures did not clear a gradient-boosted baseline. Each of those redirects effort that would otherwise be spent, and each is reproducible from the committed code.

We make no claim to outperform existing operational systems. We have not benchmarked against them, and the comparison the field usually reports is the one our own control calls into question.

\subsection{Next steps}

\emph{Track the watchlist forward.} The ranking is a prediction about lakes that have not yet failed, so it can be scored honestly only against events that have not yet happened. Freezing the current list, publishing it, and checking it against the next several seasons converts a stated skill into a measured one. Nothing else in this programme would do more to establish whether the number holds.

\emph{Re-examine the published susceptibility literature under matching.} Our control is straightforward to apply to any existing model that reports a score against an unmatched background. Repeating those evaluations with matched comparison sites would show how much of the field's reported skill is geography. We expect the effect to be substantial, since it was substantial here, and we would rather see that tested than assumed.

\emph{Pursue the landslide result properly.} Three architectures beat the baseline on landslides, and none did so decisively. That pattern is what a real but underpowered effect looks like. Landslides are the one hazard with enough catalogued events to settle the question, so the experiment is to hold the protocol fixed, increase the event count substantially, and see whether the ordering survives.

\emph{Close the gaps the limits section names.} Lake volume is the feature the literature ranks highest and the one we cannot see, so bathymetry, or a volume calibrated against it, is the single most valuable addition. Local weather and lake telemetry would lift the triggering ceiling. A full small-baseline inversion would sharpen the deformation channel that the light chain here left blunt. Finer imagery may reach the small lakes that free static data cannot separate.

\emph{Build the operational layer.} The pieces that survived are enough to schedule: refresh each lake's terrain prior and its weather state, emit a pre-season watchlist and a daily trigger level, and carry the evidence with every alert. It should feed the in-situ last mile of lake telemetry and downstream seismic alarms rather than replace it [26, 15].

\emph{Test whether the method travels.} Nothing in the design is specific to High Mountain Asia. The Andes and the Alps have their own catalogues and their own uneven coverage, and running the same protocol there would show whether the susceptibility signal and the catalogue artifact are general features of this problem.

What this paper adds is a measurement rather than a promise: how far free satellite data actually goes on each of these questions, and a watchlist and method that state their own uncertainty.

\vspace{8pt}
\noindent\rule{\linewidth}{0.4pt}
\section{References}
{\small
\setlength{\parskip}{0pt}
\begin{enumerate}[label={[\arabic*]},leftmargin=2.4em,labelsep=0.5em,itemsep=0.35em,topsep=0.4em]
\item Alaska Satellite Facility. (n.d.). \emph{HyP3: On-demand Sentinel-1 InSAR processing}. \url{https://hyp3-docs.asf.alaska.edu/}
\item Berardino, P., Fornaro, G., Lanari, R., \& Sansosti, E. (2002). A new algorithm for surface deformation monitoring based on small baseline differential SAR interferograms. \emph{IEEE Transactions on Geoscience and Remote Sensing, 40}(11), 2375--2383. \url{https://doi.org/10.1109/tgrs.2002.803792}
\item Brencher, G., Henderson, S. T., \& Shean, D. E. (2026). Quantifying degradation of the Imja Lake moraine dam with fused InSAR and SAR feature tracking time series. \emph{The Cryosphere, 20}(1), 67--86. \url{https://doi.org/10.5194/tc-20-67-2026}
\item Cook, S. J., \& Quincey, D. J. (2015). Estimating the volume of Alpine glacial lakes. \emph{Earth Surface Dynamics, 3}(4), 559--575. \url{https://doi.org/10.5194/esurf-3-559-2015}
\item Drusch, M., Del Bello, U., Carlier, S., Colin, O., Fernandez, V., Gascon, F., Hoersch, B., Isola, C., Laberinti, P., Martimort, P., Meygret, A., Spoto, F., Sy, O., Marchese, F., \& Bargellini, P. (2012). Sentinel-2: ESA's Optical High-Resolution Mission for GMES Operational Services. \emph{Remote Sensing of Environment, 120}, 25--36. \url{https://doi.org/10.1016/j.rse.2011.11.026}
\item Elkan, C., \& Noto, K. (2008). Learning classifiers from only positive and unlabeled data. \emph{Proceedings of the 14th ACM SIGKDD international conference on Knowledge discovery and data mining}, 213--220. \url{https://doi.org/10.1145/1401890.1401920}
\item European Space Agency \& Airbus. (2020). \emph{Copernicus DEM: GLO-30 global digital surface model} [Product handbook]. \url{https://spacedata.copernicus.eu/collections/copernicus-digital-elevation-model}
\item Funk, C., Peterson, P., Landsfeld, M., Pedreros, D., Verdin, J., Shukla, S., Husak, G., Rowland, J., Harrison, L., Hoell, A., \& Michaelsen, J. (2015). The climate hazards infrared precipitation with stations -- a new environmental record for monitoring extremes. \emph{Scientific Data, 2}(1), Article 150066. \url{https://doi.org/10.1038/sdata.2015.66}
\item Gao, J., Du, J., Bai, Y., Chen, T., \& Zhuoma, Y. (2024). The Impact of Climate Change on Glacial Lake Outburst Floods. \emph{Water, 16}(12), 1742. \url{https://doi.org/10.3390/w16121742}
\item GAPHAZ. (2017). \emph{Assessment of glacier and permafrost hazards in mountain regions: Technical guidance document}. Standing Group on Glacier and Permafrost Hazards in Mountains, IACS and IPA.
\item Harrison, S., Kargel, J. S., Huggel, C., Reynolds, J., Shugar, D. H., Betts, R. A., Emmer, A., Glasser, N., Haritashya, U. K., Klime\v{s}, J., Reinhardt, L., Schaub, Y., Wiltshire, A., Regmi, D., \& Vil\'{i}mek, V. (2018). Climate change and the global pattern of moraine-dammed glacial lake outburst floods. \emph{The Cryosphere, 12}(4), 1195--1209. \url{https://doi.org/10.5194/tc-12-1195-2018}
\item Huffman, G. J., Bolvin, D. T., Braithwaite, D., Hsu, K. L., Joyce, R. J., Kidd, C., Nelkin, E. J., Sorooshian, S., Stocker, E. F., Tan, J., Wolff, D. B., \& Xie, P. (2020). Integrated Multi-satellite Retrievals for the Global Precipitation Measurement (GPM) Mission (IMERG). \emph{Advances in Global Change Research}, 343--353. \url{https://doi.org/10.1007/978-3-030-24568-9_19}
\item Liu, H., Wang, Z., Wen, H., Pei, N., Xia, Z., Bian, R., Ma, S., \& Tao, L. (2025). Predicting glacial lake outburst susceptibility on the southern Tibetan Plateau with historical events and machine learning methods. \emph{Natural Hazards, 121}(15), 17677--17705. \url{https://doi.org/10.1007/s11069-025-07486-8}
\item Maclure, M. (1991). The Case-Crossover Design: A Method for Studying Transient Effects on the Risk of Acute Events. \emph{American Journal of Epidemiology, 133}(2), 144--153. \url{https://doi.org/10.1093/oxfordjournals.aje.a115853}
\item Maurer, J. M., Schaefer, J. M., Russell, J. B., Rupper, S., Wangdi, N., Putnam, A. E., \& Young, N. (2020). Seismic observations, numerical modeling, and geomorphic analysis of a glacier lake outburst flood in the Himalayas. \emph{Science Advances, 6}(38), Article eaba3645. \url{https://doi.org/10.1126/sciadv.aba3645}
\item McFEETERS, S. K. (1996). The use of the Normalized Difference Water Index (NDWI) in the delineation of open water features. \emph{International Journal of Remote Sensing, 17}(7), 1425--1432. \url{https://doi.org/10.1080/01431169608948714}
\item Morishita, Y., Lazecky, M., Wright, T., Weiss, J., Elliott, J., \& Hooper, A. (2020). LiCSBAS: An Open-Source InSAR Time Series Analysis Package Integrated with the LiCSAR Automated Sentinel-1 InSAR Processor. \emph{Remote Sensing, 12}(3), 424. \url{https://doi.org/10.3390/rs12030424}
\item Mu\~{n}oz-Sabater, J., Dutra, E., Agust\'{i}-Panareda, A., Albergel, C., Arduini, G., Balsamo, G., Boussetta, S., Choulga, M., Harrigan, S., Hersbach, H., Martens, B., Miralles, D. G., Piles, M., Rodr\'{i}guez-Fern\'{a}ndez, N. J., Zsoter, E., Buontempo, C., \& Th\'{e}paut, J. N. (2021). ERA5-Land: a state-of-the-art global reanalysis dataset for land applications. \emph{Earth System Science Data, 13}(9), 4349--4383. \url{https://doi.org/10.5194/essd-13-4349-2021}
\item Otsu, N. (1979). A Threshold Selection Method from Gray-Level Histograms. \emph{IEEE Transactions on Systems, Man, and Cybernetics, 9}(1), 62--66. \url{https://doi.org/10.1109/tsmc.1979.4310076}
\item Sattar, A., Cook, K. L., Rai, S. K., Berthier, E., Allen, S., Rinzin, S., de Vries, M. V. W., Haeberli, W., Kushwaha, P., Shugar, D. H., Emmer, A., Haritashya, U. K., Frey, H., Rao, P., Gurudin, K. S. K., Rai, P., Rajak, R., Hossain, F., Huggel, C., \ldots{} Bhat, S. Y. (2025). The Sikkim flood of October 2023: Drivers, causes, and impacts of a multihazard cascade. \emph{Science, 387}(6740), Article eads2659. \url{https://doi.org/10.1126/science.ads2659}
\item Shrestha, F., Steiner, J. F., Shrestha, R., Dhungel, Y., Joshi, S. P., Inglis, S., Ashraf, A., Wali, S., Walizada, K. M., \& Zhang, T. (2023). A comprehensive and version-controlled database of glacial lake outburst floods in High Mountain Asia. \emph{Earth System Science Data, 15}(9), 3941--3961. \url{https://doi.org/10.5194/essd-15-3941-2023}
\item Shugar, D. H., Burr, A., Haritashya, U. K., Kargel, J. S., Watson, C. S., Kennedy, M. C., Bevington, A. R., Betts, R. A., Harrison, S., \& Strattman, K. (2020). Rapid worldwide growth of glacial lakes since 1990. \emph{Nature Climate Change, 10}(10), 939--945. \url{https://doi.org/10.1038/s41558-020-0855-4}
\item Torres, R., Snoeij, P., Geudtner, D., Bibby, D., Davidson, M., Attema, E., Potin, P., Rommen, B., Floury, N., Brown, M., Traver, I. N., Deghaye, P., Duesmann, B., Rosich, B., Miranda, N., Bruno, C., L'Abbate, M., Croci, R., Pietropaolo, A., \ldots{} Rostan, F. (2012). GMES Sentinel-1 mission. \emph{Remote Sensing of Environment, 120}, 9--24. \url{https://doi.org/10.1016/j.rse.2011.05.028}
\item United Nations Development Programme. (2017). \emph{Green Climate Fund approves \$36.1 million to help Nepal protect lives and livelihoods from glacial flood risks} [Press release].
\item Veh, G., Korup, O., von Specht, S., Roessner, S., \& Walz, A. (2019). Unchanged frequency of moraine-dammed glacial lake outburst floods in the Himalaya. \emph{Nature Climate Change, 9}(5), 379--383. \url{https://doi.org/10.1038/s41558-019-0437-5}
\item Wang, W., Zhang, T., Yao, T., \& An, B. (2022). Monitoring and early warning system of Cirenmaco glacial lake in the central Himalayas. \emph{International Journal of Disaster Risk Reduction, 73}, 102914. \url{https://doi.org/10.1016/j.ijdrr.2022.102914}
\item Xu, H. (2006). Modification of normalised difference water index (NDWI) to enhance open water features in remotely sensed imagery. \emph{International Journal of Remote Sensing, 27}(14), 3025--3033. \url{https://doi.org/10.1080/01431160600589179}
\item Yang, L., Lu, Z., Zhao, C., Zhang, Q., Hu, X., \& Wang, B. (2025). Triggering factors and flooding processes of glacial lake outburst flood at Ranzerio lake. \emph{npj Natural Hazards, 2}(1), Article 90. \url{https://doi.org/10.1038/s44304-025-00147-7}
\item Yu, Y., Li, B., Li, Y., \& Jiang, W. (2024). Retrospective Analysis of Glacial Lake Outburst Flood (GLOF) Using AI Earth InSAR and Optical Images: A Case Study of South Lhonak Lake, Sikkim. \emph{Remote Sensing, 16}(13), 2307. \url{https://doi.org/10.3390/rs16132307}
\item Zhang, T., Wang, W., \& An, B. (2024). A massive lateral moraine collapse triggered the 2023 South Lhonak Lake outburst flood, Sikkim Himalayas. \emph{Landslides, 22}(2), 299--311. \url{https://doi.org/10.1007/s10346-024-02358-x}

\vspace{0.35em}
\item Lundberg, S. M., \& Lee, S. I. (2017). A unified approach to interpreting model predictions. \emph{Advances in Neural Information Processing Systems, 30}, 4765--4774.
\item Friedman, J. H. (2001). Greedy function approximation: A gradient boosting machine. \emph{The Annals of Statistics, 29}(5), 1189--1232. \url{https://doi.org/10.1214/aos/1013203451}
\item Kirschbaum, D. B., Stanley, T., \& Zhou, Y. (2015). Spatial and temporal analysis of a global landslide catalog. \emph{Geomorphology, 249}, 4--15. \url{https://doi.org/10.1016/j.geomorph.2015.03.016}
\item Roberts, D. R., Bahn, V., Ciuti, S., Boyce, M. S., Elith, J., Guillera-Arroita, G., Hauenstein, S., Lahoz-Monfort, J. J., Schr\"{o}der, B., Thuiller, W., Warton, D. I., Wintle, B. A., Hartig, F., \& Dormann, C. F. (2017). Cross-validation strategies for data with temporal, spatial, hierarchical, or phylogenetic structure. \emph{Ecography, 40}(8), 913--929. \url{https://doi.org/10.1111/ecog.02881}
\item Pekel, J. F., Cottam, A., Gorelick, N., \& Belward, A. S. (2016). High-resolution mapping of global surface water and its long-term changes. \emph{Nature, 540}(7633), 418--422. \url{https://doi.org/10.1038/nature20584}

\item Kahn, M. (2026). \emph{Toward Satellite Early Detection of Glacial Lake Outburst Floods in the Nepal Himalaya: A Feasibility Assessment Using Free Satellite Observations}. Unpublished manuscript.

\end{enumerate}
}

\vspace{4pt}
{\fontsize{8.2}{10}\selectfont\emph{Data and code.} HMAGLOFDB (doi:10.5281/zenodo.18257243); ICIMOD PDGL inventory (doi:10.26066/RDS.1971950); Sentinel-1/2, CHIRPS, GPM IMERG, ERA5-Land, and Copernicus GLO-30 via Google Earth Engine; Sentinel-1 InSAR via ASF HyP3 (\texttt{hyp3\_sdk}). All analysis code, per-milestone reports, and intermediate data are version-controlled; every figure is generated from the study's committed data.\par}

\end{document}